\documentclass[pdflatex,sn-basic]{sn-jnl}
\usepackage{graphicx}%
\usepackage{amsmath,amssymb,amsfonts}%
\usepackage[title]{appendix}%

\usepackage{comment}  

\usepackage{algorithmic}  
\usepackage{algorithm}  
\usepackage{bm}  
\usepackage{booktabs}  
\usepackage{capt-of} 

\usepackage{calc}     
\usepackage{enumitem} 
\usepackage{eurosym}  

\usepackage{graphicx} 
\usepackage{makecell}  
\usepackage{mathtools}  
\usepackage{multirow}  
\usepackage{subcaption}  
\usepackage{tabularx}  
\usepackage{threeparttable}  
\usepackage{url}  
\newcolumntype{Y}{>{\raggedright\arraybackslash}X}  

\theoremstyle{thmstyleone}%

\theoremstyle{thmstyletwo}%

\theoremstyle{thmstylethree}%

\begin{document}

\title[Article Title]{Buffered AUC maximization for scoring systems \\ via mixed-integer optimization}

\author[1]{\fnm{Moe} \sur{Shiina}}\email{s2220423@u.tsukuba.ac.jp}

\author*[2]{\fnm{Shunnosuke} \sur{Ikeda}}\email{ikeda@cs.tsukuba.ac.jp}

\author[2,3]{\fnm{Yuichi} \sur{Takano}}\email{ytakano@sk.tsukuba.ac.jp}

\affil[1]{\orgdiv{Graduate School of Science and Technology}, \orgname{University of Tsukuba}, \orgaddress{\street{1--1--1 Tennodai}, \city{Tsukuba-shi}, \postcode{305--8573}, \state{Ibaraki}, \country{Japan}}}

\affil[2]{\orgdiv{Institute of Systems and Information Engineering}, \orgname{University of Tsukuba}, \orgaddress{\street{1--1--1 Tennodai}, \city{Tsukuba-shi}, \postcode{305--8573}, \state{Ibaraki}, \country{Japan}}}

\affil[3]{\orgdiv{Center for Artificial Intelligence Research, Tsukuba Institute for Advanced Research (TIAR)}, \orgname{University of Tsukuba}, \orgaddress{\street{1--1--1 Tennodai}, \city{Tsukuba-shi}, \postcode{305--8577}, \state{Ibaraki}, \country{Japan}}}

\abstract{
A scoring system is a linear classifier composed of a small number of explanatory variables, each assigned a small integer coefficient. 
This system is highly interpretable and allows predictions to be made with simple manual calculations without the need for a calculator.
Several previous studies have used mixed-integer optimization (MIO) techniques to develop scoring systems for binary classification; however, they have not focused on directly maximizing AUC (i.e., area under the receiver operating characteristic curve), even though AUC is recognized as an essential evaluation metric for scoring systems. 
Our goal herein is to establish an effective MIO framework for constructing scoring systems that directly maximize the buffered AUC (bAUC) as the tightest concave lower bound on AUC. 
Our optimization model is formulated as a mixed-integer linear optimization (MILO) problem that maximizes bAUC subject to a group sparsity constraint for limiting the number of questions in the scoring system.  
Computational experiments using publicly available real-world datasets demonstrate that our MILO method can build scoring systems with superior AUC values compared to the baseline methods based on regularization and stepwise regression.
This research contributes to the advancement of MIO techniques for developing highly interpretable classification models.
}


\maketitle

\section{Introduction}

\subsection{Background}

In recent years, machine learning technology has been increasingly used for analytical and predictive applications in various sectors of society due to its ability to learn from available data and make intelligent decisions~\citep{sarker2021machine}.
Meanwhile, many machine learning models, typified by deep learning, are black boxes, meaning that the basis for their predictions cannot be explained in a way that humans can understand. 
The lack of transparency and accountability in such black-box models can have serious consequences, especially for high-stakes decision-making in healthcare and criminal justice, which deeply involve human lives~\citep{rudin2018optimized}. 
Since black-box models can also lead to unreliable and misleading explanations, it is preferable to use interpretable models instead for high-stakes decision-making~\citep{rudin2019stop}.

\emph{Scoring systems} are linear classification models that consist of a small number of explanatory variables with small integer coefficients~\citep{rudin2018optimized}. 
These models are highly interpretable and allow quick predictions to be made by adding and subtracting a few small integers, without the need for a calculator. 
Scoring systems are widely used in domains such as medicine, criminal justice, and finance due to their ease of use and understanding~\citep{rudin2018optimized}.
Many scoring systems have been created manually by panels of experts without using data (e.g., SAPS II score~\citep{le1993new}). 
As outlined in \cite{antman2000timi}, some scoring systems were built by rounding the coefficients of logistic regression models to integers, such as the $\mathrm{CHADS}_2$ score~\citep{gage2001validation} (Table~\ref{tab:CHADS}).   
However, these heuristic approaches often fail to satisfy the operational constraints required in practice and cannot guarantee the best possible predictive accuracy of the resultant classification models. 

\emph{Mixed-integer optimization (MIO)} is the technique for solving mathematical optimization problems involving real and integer decision variables~\citep{wolsey1999integer}. 
With advances in optimization algorithms and computer hardware~\citep{bixby2012brief,koch2022progress}, the MIO approach has recently moved into the spotlight as a high-performance model-building technique for a variety of statistical and machine learning tasks~\citep{hastie2020best,gambella2021optimization,justin2025mixed}. 
Compared to general-purpose heuristic algorithms, the MIO approach has the advantage of being able to select the best model with respect to a given objective function~\citep{miyashiro2015mixed,miyashiro2015subset,mazumder2017discrete,park2020subset,takano2020best}, subject to required constraints~\citep{bertsimas2016or,tamura2017best,tamura2019mixed,bertsimas2020scalable,chung2020mathematical,kikuchi2025two}. 
MIO methods have been successfully applied to building a variety of classification models, such as logistic regression~\citep{sato2016feature,bertsimas2017logistic,ahari2023mixed}, decision trees~\citep{bertsimas2017optimal,carrizosa2021mathematical}, and support vector machines~\citep{maldonado2014feature,tamura2024mixed}.

\begin{table*}[tbh]
\centering
\caption{The CHADS$_2$ score to assess stroke risk~\citep{gage2001validation}}
\label{tab:CHADS}
\begin{tabular}{ll}
\toprule
1. \textbf{C}ongestive Heart Failure & $+1$ point\\
2. \textbf{H}ypertension             & $+1$ point \\
3. \textbf{A}ge $\geq 75$            & $+1$ point \\
4. \textbf{D}iabetes Mellitus        & $+1$ point \\
5. Prior \textbf{S}troke or Transient Ischemic Attack & $+2$ points\\
\midrule
 \multicolumn{1}{r}{$\textbf{SCORE}=$}\\
\bottomrule
\end{tabular}

\vspace{0.8em}

\begin{tabular}{cccccccc}
\toprule
\textbf{SCORE} & 0 & 1 & 2 & 3 & 4 & 5 & 6 \\
\midrule
\textbf{RISK} & 1.9\% & 2.8\% & 4.0\% & 5.9\% & 8.5\% & 12.5\% & 18.2\% \\
\bottomrule
\end{tabular}
\end{table*}

\subsection{Related work}

MIO techniques have been used in several previous studies to develop scoring systems for binary classification. 
\cite{ustun2016supersparse} proposed the supersparse linear integer model (SLIM) for scoring systems, which directly minimizes the number of misclassifications by restricting the coefficients of selected explanatory variables to small integers.
\cite{ustun2019learning} devised the risk-calibrated supersparse linear integer model (RiskSLIM) for scoring systems, which allows risk probabilities to be calculated by minimizing the logistic loss function.
\cite{zhang2021learning} used the SLIM method to create optimal predictive checklists, where unit weights (i.e., $\pm 1$) are assigned to Boolean conditions. 
These MIO-based scoring systems have been applied to predicting recidivism rates among released prisoners~\citep{zeng2017interpretable} and mortality rates among hospitalized patients~\citep{yamga2023optimized}. 

AUC, area under the ROC (receiver operating characteristic) curve~\citep{hanley1982meaning}, has been widely used for evaluating the predictive performance of binary classification models. 
AUC is known to be more informative than the accuracy (i.e., correct classification rate) especially for imbalanced data, and has been demonstrated to be the most consistent metric across different datasets~\citep{bradley1997use,ling2003auc,li2024area}. 
It is also known that maximizing the accuracy does not necessarily maximize AUC~\citep{cortes2003auc}.
Furthermore, direct AUC maximization corresponds to an extremely difficult optimization problem of maximizing a nonconvex and discontinuous objective function~\citep{norton2019maximization}.

Various algorithms for AUC maximization have been proposed, which can be classified into four categories~\citep{yang2022auc}: linear/quadratic optimization, gradient descent, cutting-plane algorithms, and boosting methods. 
However, most of these algorithms lack theoretical validity or require iterative calculations, which significantly deteriorates the solution quality and computational efficiency of MIO models for scoring systems.

For this reason, we focus specifically on maximizing the buffered AUC (bAUC), proposed by \cite{norton2019maximization} for binary classification. 
bAUC is defined by the buffered probability of exceedance~\citep{mafusalov2018buffered} and is known as the tightest concave lower bound on AUC~\citep{norton2019maximization}. 
Moreover, bAUC maximization for linear classifiers can be reduced to linear or convex optimization by exploiting positive homogeneity~\citep{norton2019maximization}. 
\cite{tanaka2023ellipsoidal} recently applied bAUC maximization to selecting a subset of explanatory variables for ellipsoidal classifiers. 
To the best of our knowledge, however, despite AUC being recognized as an essential evaluation metric for scoring systems~\citep{zeng2017interpretable,rudin2018optimized,ustun2019learning,yamga2023optimized}, no previous studies have focused on directly maximizing AUC of scoring systems. 

\subsection*{Contribution}
Our goal herein is to establish an effective MIO framework for constructing scoring systems that directly maximize bAUC. 
Our method builds on the RiskSLIM formulation~\citep{ustun2019learning}, which trains logistic regression models with small integer regression coefficients.
First, we incorporate the concept of group sparsity~\citep{yuan2006model,bertsimas2016or} to limit the number of questions, rather than the number of response choices, in scoring systems.
Next, we adopt bAUC~\citep{norton2019maximization} as the objective function instead of the common logistic loss function.
We then transform this optimization model into an equivalent mixed-integer linear optimization (MILO) problem, which can be handled by standard MIO solvers. 

We conducted computational experiments on binary classification using publicly available real-world datasets to evaluate the effectiveness of our MILO method in comparison with baseline methods based on regularization and stepwise regression.
Our method often built scoring systems with superior AUC values compared to the baseline methods.
Although our method was generally slower than the baseline methods, its computation time was at most a few minutes, which is within a practically acceptable range for developing scoring systems.
We also validated our method by analyzing the generated score tables based on domain knowledge. 

\section{Methods}

In this section, we first introduce the existing RiskSLIM formulation~\citep{ustun2019learning} based on logistic regression for scoring systems.
Next, we define a group sparsity constraint to limit the number of questions used in scoring systems, and formulate bAUC~\citep{norton2019maximization} as a surrogate function for maximizing AUC. 
Finally, we present our scoring system optimization model and its MILO reformulation. 
Throughout this paper, we denote the set of consecutive positive integers as $[n] \coloneqq \{1, 2, \ldots, n\}$.

\subsection{RiskSLIM formulation}
Let $\{(y_i, \bm{x}_i) \mid i \in [n]\}$ be a dataset consisting of $n$ instances, where $y_i \in \{-1, +1\}$ is the binary class label to be predicted for instance $i \in [n]$, and $\bm{x}_i \coloneqq (x_{ij})_{j \in [p]} \in \{0,1\}^p$ is a vector of $p$ binary explanatory variables (yes/no responses) associated with instance $i \in [n]$.
In the logistic regression model, the conditional probability of observing class label $y \in \{-1, +1\}$ given explanatory vector $\bm{x} \in \{0,1\}^p$ is expressed as
\begin{align}\label{eq:cond_prob}
    \Pr(y \mid \bm{x}) \coloneqq \frac{1}{1 + \exp\left(-y \left(\bm{w}^\top \bm{x} + w_0\right)\right)}, 
\end{align}
where $w_0 \in \mathbb{R}$ is the intercept term, and $\bm{w} \coloneqq (w_j)_{j \in [p]} \in \mathbb{R}^{p}$ is a vector of regression coefficients.

The logistic regression parameters $(w_0, \bm{w}) \in \mathbb{R} \times \mathbb{R}^{p}$ are typically estimated by maximum likelihood estimation, which amounts to minimizing the negative log-likelihood function:
\begin{align}\label{eq:log_like}
    -\log \left(\prod_{i=1}^{n} \Pr(y_i \mid \bm{x}_i)\right) 
    & = \sum_{i=1}^{n} \log( 1 + \exp( -y_i (\bm{w}^\top \bm{x}_i + w_0) ) ) \notag \\
    & = \sum_{i=1}^{n}f(y_i (\bm{w}^{\top}\bm{x}_i + w_0)),
\end{align}
where $f(u) \coloneqq \log(1 + \exp(-u))$ is known as the logistic loss function.
Since the second derivative of the logistic loss function is always positive, the negative log-likelihood function is smooth and strictly convex.

Let $\| \bm{w} \|_0 \coloneqq |\{j \in [p] \mid w_j \neq 0\}|$ denote the $L_0$-(pseudo-)norm that counts the number of nonzero entries of a vector. 
Then, RiskSLIM~\citep{ustun2019learning} is posed as the following mixed-integer nonlinear optimization (MINLO) problem: 
\begin{subequations}\label{eq:RiskSLIM}
\begin{align}
    \underset{{w_0,\,\bm w}}{\text{minimize}}\quad~&\frac{1}{n} \sum_{i=1}^n f(y_i (\bm{w}^{\top}\bm{x}_i + w_0)) + \lambda_0 \| \bm{w} \|_0 \label{eq:RiskSLIM_obj}\\
    \text{subject to}\quad& -M\leq w_j\leq M \qquad (j\in[p]), \label{eq:RiskSLIM_boundM}\\
     &w_0\in\mathbb{R}, \quad \bm{w}\in\mathbb{Z}^{p}, \label{eq:RiskSLIM_var}
\end{align}
\end{subequations}
where $\lambda_0 \in \mathbb{R}_{+}$ is the $L_0$-norm regularization weight parameter, and $M \in \mathbb{Z}_{+}$ is the upper-bound parameter on the absolute value of the regression coefficients.
The objective in Eq.~\eqref{eq:RiskSLIM_obj} is to minimize the sum of the negative log-likelihood function in Eq.~\eqref{eq:log_like} and the $L_0$-norm regularization term that reduces the number of explanatory variables used for prediction. 
Eqs.~\eqref{eq:RiskSLIM_boundM} and \eqref{eq:RiskSLIM_var} ensure that all regression coefficients are estimated as small integers. 

Note that the RiskSLIM problem involves a nonlinear objective function in Eq.~\eqref{eq:RiskSLIM_obj}. 
Therefore, solving the problem~\eqref{eq:RiskSLIM} requires specialized techniques such as piecewise linear approximation~\citep{sato2016feature,saishu2021sparse}, cutting-plane algorithms~\citep{bertsimas2017logistic,ustun2019learning}, or exponential cone optimization~\citep{ahari2023mixed}.

\subsection{Group sparsity}

Let $\{J_s\}_{s \in [q]}$ be a partition of the index set of explanatory variables, where $J_s \subseteq [p]$ corresponds to the group of explanatory variables associated with question $s \in [q]$.
As a typical example, let us consider gender (a categorical variable) and age (a quantitative variable), which can be converted into dummy variables as follows:
\begin{description}[leftmargin=!,labelwidth=\widthof{$\mathbf{Gender} ~(s=1):$},font=\bfseries]
    \item[$\mathbf{Gender} ~(s=1):$] $\mbox{Woman}~(j=1)$, $\mbox{Man}~(j=2)$, $\mbox{Non-binary}~(j=3)$; 
    \item[$\mathbf{Age} ~(s=2):$] $\mbox{Age} \le 20~(j=4)$, $21 \le \mbox{Age} \le 40~(j=5)$, $41 \le \mbox{Age} \le 60~(j=6)$, $61 \le \mbox{Age} \le 80~(j=7)$. 
\end{description}
In this case, the group index sets are defined as $J_1 \coloneqq \{1,2,3\}$ and $J_2 \coloneqq \{4,5,6,7\}$. 
Note that asking about gender (or age) will provide responses for all the associated explanatory variables $j \in J_1$ (or $j \in J_2$).
With this in mind, we consider group sparsity~\citep{yuan2006model,bertsimas2016or}, by limiting the number of questions rather than the number of explanatory variables (e.g., $\| \bm{w} \|_0$ in Eq.~\eqref{eq:RiskSLIM_obj}).

We introduce a binary decision variable $\bm{z} \coloneqq (z_s)_{s \in [q]} \in \{0, 1\}^{q}$ for group sparsity such that
\begin{align*}
z_s \coloneqq 
\begin{cases}
~1 & \text{if group } J_s \text{ is selected},\\
~0 & \text{otherwise}
\end{cases} \qquad (s \in [q]).
\end{align*}
The selection of questions in scoring systems can be represented as
\begin{align}
    &z_s = 0 ~\Rightarrow~ w_j = 0 \qquad (j \in J_s,~ s \in [q]), \label{eq:logical}\\
    &\sum_{s=1}^{q} z_s \leq \theta, \label{eq:cardinality}
\end{align}
where $\theta \in \mathbb{Z}_{+}$ is a parameter for specifying the maximum number of questions.
Eq.~\eqref{eq:logical} is the logical implication ensuring that if question $s \in [q]$ is unselected, all the associated explanatory variables are eliminated from the regression model by setting their coefficients to zero. 
Eq.~\eqref{eq:cardinality} is the cardinality constraint for limiting the number of questions.

\subsection{Buffered AUC maximization}
AUC stands for the two-dimensional area under the ROC curve, which plots the relationship between the false positive rate and the true positive rate at various classification thresholds.
It is also known (e.g., in \cite{hanley1982meaning}) that the empirical AUC can be calculated as the probability that a randomly selected positive instance will be scored higher than a randomly selected negative instance.

Let $I_{+} \coloneqq \{i \in [n] \mid y_i = +1\}$ and $I_{-} \coloneqq \{i \in [n] \mid y_i = -1\}$ be the index sets of positive and negative instances, respectively, and $n_{+} \coloneqq |I_{+}|$ and $n_{-} \coloneqq |I_{-}|$ be the corresponding numbers of instances. 
Then, we use the ranking error $\bm{w}^{\top}(\bm{x}_{\ell} - \bm{x}_i)$ for each positive--negative instance pair $(i,\ell) \in I_{+} \times I_{-}$ to calculate the empirical AUC:
\begin{align}
    \text{AUC}(\bm{w}) 
    & \coloneqq \frac{1}{n_{+} n_{-}} \sum_{i \in I_{+}} \sum_{\ell \in I_{-}} \mathbb{I}(\bm{w}^{\top} \bm{x}_i > \bm{w}^{\top} \bm{x}_{\ell}) \notag \\
    & = 1 - \frac{1}{n_{+} n_{-}} \sum_{i \in I_{+}} \sum_{\ell \in I_{-}} \mathbb{I}_{\mathbb{R}_{+}}(\bm{w}^{\top}(\bm{x}_{\ell} - \bm{x}_i)), \label{eq:empAUC}
\end{align}
where $\mathbb{I}_{\mathbb{R}_{+}}(x)$ is the indicator function of $\mathbb{R}_{+}$; that is, $\mathbb{I}_{\mathbb{R}_{+}}(x) = 1$ if $x \ge 0$, and $\mathbb{I}_{\mathbb{R}_{+}}(x) = 0$ otherwise. 
However, directly maximizing AUC is computationally intractable because the indicator function is nonconvex and discontinuous.

\begin{figure}[tbp] 
    \centering 
    \includegraphics[width=9cm]{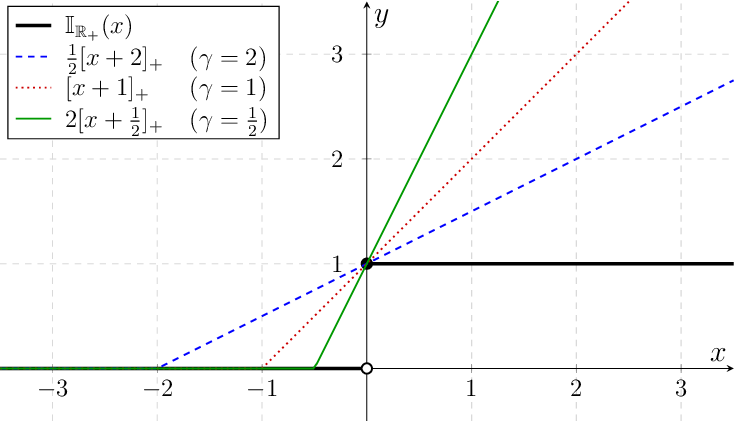}
    \caption{Relationship between $y = \mathbb{I}_{\mathbb{R}_{+}}(x)$ and $y = \frac{1}{\gamma}[x + \gamma]_{+}$ for $\gamma \in \{\frac{1}{2}, 1, 2\}$} 
    \label{fig:func}
\end{figure}

To avoid this computational challenge, \cite{norton2019maximization} proposed bAUC, which is computationally tractable and serves as the tightest concave lower bound on AUC. 
Note here that the following relationship holds for each $\gamma > 0$, as shown in Fig.~\ref{fig:func}:
\[
\mathbb{I}_{\mathbb{R}_{+}}(x) \le \frac{1}{\gamma}[x + \gamma]_{+} \qquad (x \in \mathbb{R}),
\]
where $[\,x\,]_{+} \coloneqq \max\{x,0\}$ is a function that returns the positive part of its argument. 
Applying this relationship to Eq.~\eqref{eq:empAUC} derives bAUC as follows:
\begin{align}\label{eq:bAUC_def}
    \text{bAUC}(\bm{w}) \coloneqq 1 - \min_{\gamma \in \mathbb{R}_{+}} \left\{\frac{1}{n_{+} n_{-} \gamma} \sum_{i \in I_{+}} \sum_{\ell \in I_{-}} [\bm{w}^{\top}(\bm{x}_{\ell} - \bm{x}_i) + \gamma]_{+} \right\}.
\end{align}
By following \cite{norton2019maximization}, the bAUC maximization problem can be rewritten using the variable transformation $\bm{w} \leftarrow \gamma \tilde{\bm{w}}$ as follows:
\begin{align}
\max_{\bm{w} \in \mathbb{R}^p} \text{bAUC}(\bm{w}) 
& \iff \min_{(\bm{w},\gamma) \in \mathbb{R}^p \times \mathbb{R}_{+}} \left\{\frac{1}{n_{+} n_{-} \gamma} \sum_{i \in I_{+}} \sum_{\ell \in I_{-}} [\bm{w}^{\top}(\bm{x}_{\ell} - \bm{x}_i) + \gamma]_{+} \right\} \notag \\
& \iff \min_{(\tilde{\bm{w}},\gamma) \in \mathbb{R}^p \times \mathbb{R}_{+}} \left\{\frac{1}{n_{+} n_{-} \gamma} \sum_{i \in I_{+}} \sum_{\ell \in I_{-}} [\gamma\tilde{\bm{w}}^{\top}(\bm{x}_{\ell} - \bm{x}_i) + \gamma]_{+} \right\} \notag \\
& \iff \min_{\tilde{\bm{w}} \in \mathbb{R}^p} \left\{\frac{1}{n_{+} n_{-}} \sum_{i \in I_{+}} \sum_{\ell \in I_{-}} [\tilde{\bm{w}}^{\top}(\bm{x}_{\ell} - \bm{x}_i) + 1]_{+} \right\}. \notag
\end{align}
This transformation can also be interpreted as $\gamma = 1$ being an optimal solution to the minimization problem in Eq.~\eqref{eq:bAUC_def}. 

\subsection{Our MILO formulation}

We will now formulate our scoring system optimization model. 
To remove explanatory variables with no predictive power, we introduce the $L_1$-norm regularization term $\lambda_1 \|\bm{w}\|_1 \coloneqq \lambda_1 \sum_{j=1}^p |w_j|$ to be minimized, where $\lambda_1 \in \mathbb{R}_{+}$ is the $L_1$-norm regularization weight parameter. 
Our optimization model maximizes bAUC subject to the constraints of group sparsity and small integer regression coefficients: 
\begin{subequations}\label{eq:ourForm1}
\begin{align}
    \underset{\bm{w},\,\bm{z}}{\text{minimize}} \quad & \frac{1}{n_{+} n_{-}} \sum_{i \in I_{+}} \sum_{\ell \in I_{-}} [\bm{w}^{\top}(\bm{x}_{\ell} - \bm{x}_i) + 1]_{+} + \lambda_1 \| \bm{w} \|_1 \label{eq:ourForm1_obj} \\
    \text{subject to} \quad     & z_s=0 ~\Rightarrow~ w_j=0 \qquad (j\in J_s,~s\in[q]) ,\label{eq:ourForm1_group1}\\
     &\sum_{s=1}^{q}z_s\leq\theta, \label{eq:ourForm1_group2}\\
     & -M\leq w_j\leq M \qquad (j\in[p]), \label{eq:ourForm1_boundM}\\
     &\bm{w}\in\mathbb{Z}^{p}, \quad \bm{z}\in\{0,1\}^q. \label{eq:ourForm1_var}
\end{align}
\end{subequations}
However, the objective function in Eq.~\eqref{eq:ourForm1_obj} is a non-differentiable nonlinear function, making it difficult to handle.

Let $\bm{w}^{+} \coloneqq (w^{+}_j)_{j \in [p]} \in \mathbb{R}^p_{+}$ and $\bm{w}^{-} \coloneqq (w^{-}_j)_{j \in [p]} \in \mathbb{R}^p_{+}$ be auxiliary decision variables introduced to decompose the regression coefficients into positive and negative parts; that is, $(w^{+}_j, w^{-}_j) = (w_j, 0)$ if $w_j \ge 0$, and $(w^{+}_j, w^{-}_j) = (0, -w_j)$ otherwise. 
Also, let $\bm{v} \coloneqq (v_{i\ell})_{(i,\ell) \in I_{+} \times I_{-}} \in \mathbb{R}^{n_{+} \times n_{-}}_{+}$ be an auxiliary decision variable introduced to represent the positive part $[\,\cdot\,]_{+}$. 
We can then reformulate problem~\eqref{eq:ourForm1} as the following MILO problem: 
\begin{subequations}\label{eq:ourForm2}
\begin{align}
    \underset{{\bm v},\,{\bm w},\, {\bm w^{+}},\, {\bm w^{-}},\,\bm{z}}{\text{minimize}}\quad&\frac{1}{n_{+} n_{-}}\sum_{i \in I_{+}}\sum_{\ell\in I_{-}}v_{i\ell} + \lambda_1 \sum_{j=1}^{p}(w^{+}_{j} + w^{-}_{j}) \label{eq:ourForm2_obj} \\
    \text{subject to}\quad~ & v_{i\ell} \geq \bm{w}^{\top}(\bm{x}_{\ell}-\bm{x}_{i})+1 \qquad (i\in I_{+}, ~\ell\in I_{-}), \label{eq:ourForm2_hinge}\\
     & w_j=w^{+}_{j} - w^{-}_{j} \qquad (j\in[p]), \label{eq:ourForm2_L1}\\
     & - M z_s \le w_j \le M z_s \qquad (j\in J_s,~s\in[q]),\label{eq:ourForm2_group1}\\
     &\sum_{s=1}^{q}z_s\leq\theta, \label{eq:ourForm2_group2}\\
     &\bm{v}\in\mathbb{R}_+^{n_{+}\times n_{-}}, \quad \bm{w}\in\mathbb{Z}^{p}, \quad \bm{w}^{+},\bm{w}^{-}\in\mathbb{R}_+^{p}, \quad \bm{z}\in\{0,1\}^q. \label{eq:ourForm2_var}
\end{align}
\end{subequations}
The first term in Eq.~\eqref{eq:ourForm2_obj} is combined with Eq.~\eqref{eq:ourForm2_hinge} to calculate the empirical bAUC, and the second term in Eq.~\eqref{eq:ourForm2_obj} is combined with Eq.~\eqref{eq:ourForm2_L1} to calculate the $L_1$-regularization term. 
Eq.~\eqref{eq:ourForm2_group1} is a linear constraint equivalent to the logical implication in Eq.~\eqref{eq:logical}, which together with Eq.~\eqref{eq:ourForm2_group2} limits the number of questions through group sparsity. 
Note that Eqs.~\eqref{eq:ourForm2_group1} and \eqref{eq:ourForm2_var} ensure that the regression coefficients are small integers. 

In contrast to the existing MINLO problem~\eqref{eq:RiskSLIM} that requires specialized solution techniques, our MILO problem~\eqref{eq:ourForm2} can be solved directly with standard MIO solvers.
After solving the problem~\eqref{eq:ourForm2}, we can calculate the conditional probability in Eq.~\eqref{eq:cond_prob} by maximum likelihood estimation of the intercept term $w_0 \in \mathbb{R}$ while fixing the obtained regression coefficients $\bm{w} \in \mathbb{Z}^p$. 

\section{Experimental results and discussion}
In this section, we evaluate the effectiveness of our scoring system optimization model through numerical experiments using publicly available real-world datasets.
All computations were performed on a macOS 12.6 computer equipped with an Apple M1 chip (8 cores) and 8 GB of RAM.

\subsection{Datasets}
We conducted comprehensive experiments using four real-world datasets for binary classification downloaded from the UC Irvine Machine Learning Repository~\citep{uci_repository}. 
Table~\ref{tab:datasets} lists the number of instances ($n$), the proportion of positive instances ($n_{+}/n$), the number of binary explanatory variables ($p$), and the number of groups for explanatory variables ($q$) for each dataset.

All explanatory variables were converted into dummy variables (yes/no responses) for use in scoring systems.
Specifically, each quantitative (numeric) variable was converted into three binary variables by setting two thresholds that divided the number of data instances into approximately equal thirds.
Each qualitative (categorical) variable (with more than two levels) was encoded into up to four dummy variables: three corresponding to the most frequent categories and one aggregating all remaining categories (if present). 
Note that in the $\mathtt{bank}$ dataset, the variable ``duration'' was excluded because it is not suitable for predictive purposes.

To evaluate predictive accuracy, we randomly split each dataset into a training set (80\%) and a testing set (20\%).
We repeated this random split five times and report the average of the results from these five independent runs.
Standard errors are shown as error bars in figures and in parentheses in tables.

\begin{table}[t]
    \centering
    \caption{Summary of the datasets}
    \label{tab:datasets}
    \begin{tabularx}{\textwidth}{lrrrrY}
        \toprule
        Abbreviation & \multicolumn{1}{c}{$n$} & \multicolumn{1}{c}{$n_{+}/n$} & \multicolumn{1}{c}{$p$} & \multicolumn{1}{c}{$q$} & Task (Dataset name~\citep{uci_repository}) \\ \midrule
        $\mathtt{surgery}$
        & 470 & 14.9$\%$ & 34 & 16 & Predict whether a lung cancer patient will die within one year after surgery (Thoracic Surgery Data) \\
        $\mathtt{mushroom}$
        & 8124 & 46.1$\%$ & 68 & 21 & Predict whether a mushroom is poisonous (Mushroom) \\  
        $\mathtt{bank}$
        & 45211 & 11.3$\%$ & 46 & 16 & Predict whether a client will subscribe a term deposit (Bank Marketing) \\
        $\mathtt{adult}$
        & 48842 & 22.8$\%$ & 47 & 14 & Predict whether an individual's annual income exceeds \$50,000 (Adult) \\
        \bottomrule
    \end{tabularx}
\end{table}

\subsection{Baseline methods}
To evaluate the effectiveness of our method, we compared its performance with the following benchmark methods:
\begin{description}[leftmargin=!,labelwidth=\widthof{\bf{Backward-Elimination:}}]
    \item[$\mathbf{bAUC\text{-}Integer}$] Our method that solves the MILO problem~\eqref{eq:ourForm2} for bAUC maximization.
    \vspace{2mm}
    \item[$\mathbf{bAUC\text{-}Rounding}$] Benchmark method that solves the MILO problem~\eqref{eq:ourForm2} without the integer constraint $\bm{w}\in\mathbb{Z}^{p}$. 
    This method can be regarded as the addition of the group sparsity constraints in Eqs.~\eqref{eq:logical} and \eqref{eq:cardinality} to the existing subset selection method~\citep{tanaka2023ellipsoidal} for bAUC maximization. 
    \vspace{2mm}
    \item[$L_1\text{-}\mathbf{Regularization}$] Logistic regression model with $L_1$-regularization~\citep{friedman2010regularization}. 
    We traversed a regularization path by varying the regularization weight parameter and selected the most strongly regularized model with non-zero regression coefficients for approximately $\theta$ groups of explanatory variables.
    \vspace{2mm}
    \item[$\mathbf{Elastic\text{-}Net}$] Logistic regression model with elastic-net regularization~\citep{friedman2010regularization}. 
    As in the case of the $L_1$-Regularization, the regularization weight parameter was tuned to select a model with approximately $\theta$ groups of explanatory variables.
    \vspace{2mm}
    \item[$\mathbf{Forward\text{-}Selection}$] Forward stepwise selection method for logistic regression~\citep{ferri1994comparative}. 
    In this method, explanatory variables with the largest increase in AUC on a validation set (20\% of the training set) were added one by one to the null model until the number of selected groups of explanatory variables reached $\theta$.
    \vspace{2mm}
    \item[$\mathbf{Backward\text{-}Elimination}$] Backward stepwise elimination method for logistic regression~\citep{ferri1994comparative}. 
    In this method, explanatory variables with the smallest decrease in AUC on a validation set (20\% of the training set) were removed one by one from the full model, as long as $\theta$ groups of explanatory variables were maintained.
    
\end{description}

Since all the baseline methods (except our bAUC-Integer method) generate continuous regression coefficients, we applied a post-processing step based on \cite{ustun2019learning} to convert these continuous coefficients to integers.
Specifically, the obtained continuous coefficients $\bm{w} \in \mathbb{R}^p$ were scaled and rounded to the nearest integers $\bm{w} \in \mathbb{Z}^p$ as
\begin{align*}
    w_j \leftarrow \operatorname{round}\left( \frac{w_j}{\max_{k \in [p]} |w_k|} (M + 0.49) \right) \qquad (j \in [p]),
\end{align*}
where $\operatorname{round}(\cdot)$ is a function that rounds its argument to the nearest integer.

The MILO problems were solved using Gurobi Optimizer\footnote{\url{https://www.gurobi.com/}} 12.0.1.
For both the bAUC-Integer and bAUC-Rounding methods, the regularization weight parameter was fixed at $\lambda_1 = 0.005$.
To mitigate the high computational complexity of these methods, we randomly sampled 300 instances from the training set to solve the MILO problems.
The other baseline methods were implemented using the Python \texttt{scikit-learn} library~\citep{pedregosa2011scikit}.
For the Elastic-Net method, the mixing parameter of the $L_1$- and $L_2$-regularization terms was set to $0.5$.

\begin{table*}[!tbh]
    \centering
    \caption{Testing AUC values across methods for building scoring systems}
    \label{tab:auc}
    \setlength{\tabcolsep}{2pt}  
    \small  
    \begin{subtable}{\textwidth}
        \centering
        \caption{$\mathtt{surgery}$ dataset} 
        \label{tab:auc_surgery}
        \begin{tabular}{lrrrrrrr}
        \toprule
        \multirow{2}{*}[-0.5ex]{Method}
            & \multicolumn{3}{c}{$M=1$}
            & \multicolumn{3}{c}{$M=2$}
            & \multirow{2}{*}[-0.5ex]{Average} \\
        \cmidrule(lr){2-4} \cmidrule(lr){5-7}
            & \multicolumn{1}{c}{$\theta = 4$} & \multicolumn{1}{c}{$\theta = 6$} & \multicolumn{1}{c}{$\theta = 8$}
            & \multicolumn{1}{c}{$\theta = 4$} & \multicolumn{1}{c}{$\theta = 6$} & \multicolumn{1}{c}{$\theta = 8$} & \\
        \midrule
        bAUC-Integer & 0.603 & \textbf{0.627} & \textbf{0.628} & 0.603 & \textbf{0.627} & 0.624 & \textbf{0.619} \\
        &($\pm$0.016)&($\pm$0.032)&($\pm$0.036)&($\pm$0.014)&($\pm$0.029)&($\pm$0.039)&\\
        \addlinespace[2mm]
        bAUC-Rounding                  & \textbf{0.609} & 0.620 & 0.611   & \textbf{0.607} & 0.622 & 0.629 & 0.616 \\
        &($\pm$0.021)&($\pm$0.034)&($\pm$0.034)&($\pm$0.016)&($\pm$0.016)&($\pm$0.024)&\\
        \addlinespace[2mm]
        $L_1$-Regularization  & 0.565 & 0.578          & 0.605   & 0.603 & 0.609 & 0.606           & 0.594 \\
        &($\pm$0.029)&($\pm$0.033)&($\pm$0.022)&($\pm$0.021)&($\pm$0.018)&($\pm$0.023)&\\
        \addlinespace[2mm]
        Elastic-Net           & 0.539 & 0.557 & 0.565   & 0.533           & 0.550           & 0.576           & 0.553 \\
        &($\pm$0.015)&($\pm$0.008)&($\pm$0.038)&($\pm$0.029)&($\pm$0.028)&($\pm$0.023)&\\
        \addlinespace[2mm]
        Forward-Selection     & 0.551 & 0.548 & 0.548   & 0.538           & 0.577           & 0.577           & 0.557 \\
        &($\pm$0.025)&($\pm$0.033)&($\pm$0.033)&($\pm$0.024)&($\pm$0.29)&($\pm$0.039)&\\
        \addlinespace[2mm]
        Backward-Elimination  & 0.560 & 0.587 & 0.602   & 0.580           & 0.600           & \textbf{0.637}           & 0.594 \\
        &($\pm$0.026)&($\pm$0.032)&($\pm$0.045)&($\pm$0.043)&($\pm$0.032)&($\pm$0.036)&\\
        \bottomrule
        \end{tabular}
    \end{subtable}

    \vspace{3mm}

    \begin{subtable}{\textwidth}
        \centering
        \caption{$\mathtt{mushroom}$ dataset} 
        \label{tab:auc_mushroom}
        \begin{tabular}{lrrrrrrr}
        \toprule
        \multirow{2}{*}[-0.5ex]{Method}
            & \multicolumn{3}{c}{$M=1$}
            & \multicolumn{3}{c}{$M=2$}
            & \multirow{2}{*}[-0.5ex]{Average} \\
        \cmidrule(lr){2-4} \cmidrule(lr){5-7}
            & \multicolumn{1}{c}{$\theta = 4$} & \multicolumn{1}{c}{$\theta = 6$} & \multicolumn{1}{c}{$\theta = 8$}
            & \multicolumn{1}{c}{$\theta = 4$} & \multicolumn{1}{c}{$\theta = 6$} & \multicolumn{1}{c}{$\theta = 8$} & \\
        \midrule
        bAUC-Integer & \textbf{0.985} & \textbf{0.989} & \textbf{0.990} & 0.985 & 0.984 & 0.988 & \textbf{0.987} \\
        &($\pm$0.002)&($\pm$0.001)&($\pm$0.002)&($\pm$0.002)&($\pm$0.002)&($\pm$0.002)&\\
        \addlinespace[2mm]
        bAUC-Rounding                  & 0.979 & 0.985 & 0.971 & 0.985 & 0.985 & 0.988 & 0.982 \\
        &($\pm$0.005)&($\pm$0.003)&($\pm$0.009)&($\pm$0.002)&($\pm$0.001)&($\pm$0.002)&\\
        \addlinespace[2mm]
        $L_1$-Regularization  & 0.947 & 0.976 & 0.976 & 0.981 & 0.981 & 0.983 & 0.974 \\
        &($\pm$0.008)&($\pm$0.001)&($\pm$0.001)&($\pm$0.002)&($\pm$0.001)&($\pm$0.001)&\\
        \addlinespace[2mm]
        Elastic-Net           & 0.889 & 0.889 & 0.969 & 0.889 & 0.941 & 0.983 & 0.927 \\
        &($\pm$0.002)&($\pm$0.002)&($\pm$0.007)&($\pm$0.002)&($\pm$0.002)&($\pm$0.001)&\\
        \addlinespace[2mm]
        Forward-Selection     & 0.970 & 0.954 & 0.961 & \textbf{0.987} & \textbf{0.990} & \textbf{0.992} & 0.976 \\
        &($\pm$0.004)&($\pm$0.010)&($\pm$0.010)&($\pm$0.002)&($\pm$0.002)&($\pm$0.002)&\\
        \addlinespace[2mm]
        Backward-Elimination  & 0.916 & 0.924 & 0.934 & 0.983 & 0.989 & \textbf{0.992} & 0.956 \\
        &($\pm$0.021)&($\pm$0.020)&($\pm$0.024)&($\pm$0.002)&($\pm$0.002)&($\pm$0.001)&\\
        \bottomrule
        \end{tabular}
    \end{subtable}
\end{table*}

\begin{table*}[!tbh]\ContinuedFloat
    \centering
    \caption{Testing AUC values across methods for building scoring systems (continued)}
    \setlength{\tabcolsep}{2pt}  
    \small  
    \begin{subtable}{\textwidth}
        \centering
        \caption{$\mathtt{bank}$ dataset} 
        \label{tab:auc_bankmarketing}
        \begin{tabular}{lrrrrrrr}
        \toprule
        \multirow{2}{*}[-0.5ex]{Method}
            & \multicolumn{3}{c}{$M=1$}
            & \multicolumn{3}{c}{$M=2$}
            & \multirow{2}{*}[-0.5ex]{Average} \\
        \cmidrule(lr){2-4} \cmidrule(lr){5-7}
            & \multicolumn{1}{c}{$\theta = 4$} & \multicolumn{1}{c}{$\theta = 6$} & \multicolumn{1}{c}{$\theta = 8$}
            & \multicolumn{1}{c}{$\theta = 4$} & \multicolumn{1}{c}{$\theta = 6$} & \multicolumn{1}{c}{$\theta = 8$} & \\
        \midrule
        bAUC-Integer & 0.630 & 0.644 & 0.632 & 0.632 & 0.628 & 0.638 & 0.634 \\
        &($\pm$0.029)&($\pm$0.017)&($\pm$0.016)&($\pm$0.024)&($\pm$0.018)&($\pm$0.015)&\\
        \addlinespace[2mm]
        bAUC-Rounding  & 0.594 & 0.596 & 0.594 & 0.636 & 0.640 & 0.645 & 0.618 \\
        &($\pm$0.023)&($\pm$0.027)&($\pm$0.026)&($\pm$0.017)&($\pm$0.016)&($\pm$0.007)&\\
        \addlinespace[2mm]
        $L_1$-Regularization  & 0.660 & 0.659 & \textbf{0.698} & 0.660 & 0.665 & \textbf{0.708} & 0.675 \\
        &($\pm$0.003)&($\pm$0.003)&($\pm$0.003)&($\pm$0.003)&($\pm$0.006)&($\pm$0.004)&\\
        \addlinespace[2mm]
        Elastic-Net & 0.500 & \textbf{0.698} & 0.697 & 0.500 & 0.698 & 0.707 & 0.633 \\
        &($\pm$0.000)&($\pm$0.005)&($\pm$0.003)&($\pm$0.000)&($\pm$0.003)&($\pm$0.003)&\\
        \addlinespace[2mm]
        Forward-Selection & 0.667 & 0.665 & 0.665 & \textbf{0.700} & \textbf{0.707} & 0.710 & \textbf{0.686} \\
        &($\pm$0.003)&($\pm$0.002)&($\pm$0.002)&($\pm$0.002)&($\pm$0.004)&($\pm$0.003)&\\
        \addlinespace[2mm]
        Backward-Elimination  & \textbf{0.671} & 0.665 & 0.665 & 0.692 & 0.698 & 0.704 & 0.683 \\
        &($\pm$0.007)&($\pm$0.002)&($\pm$0.002)&($\pm$0.007)&($\pm$0.003)&($\pm$0.001)&\\
        \bottomrule
        \end{tabular}
    \end{subtable}

    \vspace{3mm}

    \begin{subtable}{\textwidth}
        \centering
        \caption{$\mathtt{adult}$ dataset} 
        \label{tab:auc_adult}
        \begin{tabular}{lrrrrrrr}
        \toprule
        \multirow{2}{*}[-0.5ex]{Method}
            & \multicolumn{3}{c}{$M=1$}
            & \multicolumn{3}{c}{$M=2$}
            & \multirow{2}{*}[-0.5ex]{Average} \\
        \cmidrule(lr){2-4} \cmidrule(lr){5-7}
            & \multicolumn{1}{c}{$\theta = 4$} & \multicolumn{1}{c}{$\theta = 6$} & \multicolumn{1}{c}{$\theta = 8$}
            & \multicolumn{1}{c}{$\theta = 4$} & \multicolumn{1}{c}{$\theta = 6$} & \multicolumn{1}{c}{$\theta = 8$} & \\
        \midrule
        bAUC-Integer & \textbf{0.846} & \textbf{0.855} & \textbf{0.861} & 0.849 & 0.862 & 0.865 & \textbf{0.856} \\
        &($\pm$0.010)&($\pm$0.005)&($\pm$0.006)&($\pm$0.011)&($\pm$0.006)&($\pm$0.006)&\\
        \addlinespace[2mm]
        bAUC-Rounding                  & 0.792 & 0.830 & 0.846 & 0.849 & 0.857 & 0.863 & 0.840 \\
        &($\pm$0.050)&($\pm$0.006)&($\pm$0.008)&($\pm$0.009)&($\pm$0.004)&($\pm$0.008)&\\
        \addlinespace[2mm]
        $L_1$-Regularization  & 0.771 & 0.835 & 0.855 & 0.804 & \textbf{0.870} & 0.873 & 0.835 \\
        &($\pm$0.001)&($\pm$0.004)&($\pm$0.001)&($\pm$0.007)&($\pm$0.001)&($\pm$0.002)&\\
        \addlinespace[2mm]
        Elastic-Net           & 0.763 & 0.824 & 0.855 & 0.763 & 0.848 & \textbf{0.876} & 0.822 \\
        &($\pm$0.001)&($\pm$0.001)&($\pm$0.001)&($\pm$0.001)&($\pm$0.001)&($\pm$0.001)&\\
        \addlinespace[2mm]
        Forward-Selection     & 0.806 & 0.806 & 0.819 & \textbf{0.861} & 0.858 & 0.849 & 0.833 \\
        &($\pm$0.002)&($\pm$0.002)&($\pm$0.008)&($\pm$0.001)&($\pm$0.005)&($\pm$0.005)&\\
        \addlinespace[2mm]
        Backward-Elimination  & 0.809 & 0.815 & 0.828 & 0.858 & 0.855 & 0.854 & 0.837 \\
        &($\pm$0.002)&($\pm$0.007)&($\pm$0.006)&($\pm$0.004)&($\pm$0.005)&($\pm$0.003)&\\
        \bottomrule
        \end{tabular}
    \end{subtable}
\end{table*}

\subsection{Predictive accuracy}
Table \ref{tab:auc} presents the testing AUC values for each method when setting the upper bound on regression coefficients to $M \in \{1, 2\}$ and the maximum number of questions to $\theta \in \{4, 6, 8\}$. 
Note that the best AUC value is shown in bold for each parameter configuration of $(M,\theta)$. 

For the $\mathtt{surgery}$ dataset (Table \ref{tab:auc_surgery}), both the bAUC-Integer and bAUC-Rounding methods clearly outperformed the remaining baseline methods.
In terms of the average AUC value across all $(M, \theta)$ configurations, our bAUC-Integer method (0.619) slightly exceeded the bAUC-Rounding method (0.616), whereas the remaining baseline methods showed significantly lower averages below 0.594.
Specifically, the highest AUC values were achieved by our bAUC-Integer method for $(M,\theta) \in \{(1,6),(1,8),(2,6)\}$, by the bAUC-Rounding method for $(M,\theta) \in \{(1,4),(2,4)\}$, and by the Backward-Elimination method for $(M,\theta) = (2,8)$.
The standard errors of AUC for our bAUC-Integer method were comparable to those for the baseline methods, indicating the stability of its prediction accuracy.

For the $\mathtt{mushroom}$ dataset (Table \ref{tab:auc_mushroom}), all the methods achieved very high AUC values.
Even in this relatively easy classification task, our bAUC-Integer method consistently achieved the highest AUC values when $M=1$ and competitive AUC values when $M=2$, resulting in the highest average AUC value (0.987). 
In particular, our bAUC-Integer method achieved $\text{AUC}=0.985$ when $(M,\theta) = (1,4)$, demonstrating its remarkable ability to accurately identify poisonous mushrooms with manual calculations based on just four questions.
The standard errors of AUC for all methods were extremely small, indicating statistically significant differences between the methods.
These results suggest that even in easy classification tasks, directly maximizing bAUC using our integer-constrained model can provide consistent benefits over the baseline methods.

For the $\mathtt{bank}$ dataset (Table \ref{tab:auc_bankmarketing}), conventional regularization and stepwise methods outperformed the bAUC maximization methods.
Specifically, the average AUC value was higher for the Forward-Selection method (0.686), Backward-Elimination method (0.683), and $L_1$-Regularization method (0.675) than for the bAUC-Integer method (0.634), Elastic-Net method (0.633), and bAUC-Rounding method (0.618).
This result is likely due to the imbalanced $\mathtt{bank}$ dataset containing a large number of instances (Table \ref{tab:datasets}).
As mentioned above, the bAUC maximization methods sampled 300 instances to reduce computation time, which significantly reduced the training dataset size and thus potentially amplified the negative impact of class label imbalance.

For the $\mathtt{adult}$ dataset (Table \ref{tab:auc_adult}), our bAUC-Integer method again achieved the best average AUC value (0.856), outperforming the baseline methods.
Its advantage over the bAUC-Rounding method was particularly pronounced for $M=1$, where our bAUC-Integer method consistently produced higher AUC values with smaller standard errors.
When $M = 2$, several baseline methods achieved the best AUC values for particular $\theta$ values (i.e., Forward-Selection at $\theta =4$, $L_1$-Regularization at $\theta = 6$, and Elastic-Net at $\theta = 8$), but their averages fell below that of our bAUC-Integer method.
Overall, these results demonstrate that our method combines high accuracy with low variability to improve predictive performance on moderately difficult high-dimensional datasets.

In summary, our method provided the highest average AUC values for three of the four real-world datasets.
These results demonstrate that directly maximizing bAUC using MIO formulation techniques is highly effective in developing scoring systems with strong discriminatory power.

\begin{table*}[!tbh]
    \centering
    \caption{Training computation times [s] across methods for building scoring systems}
    \label{tab:computation}
    \catcode`?=\active \def?{\phantom{0}}
    \setlength{\tabcolsep}{1pt}  
    \small  
    \begin{subtable}{\textwidth}
        \centering
        \caption{$M = 2$}
        \label{tab:comp_theta}
        \begin{tabular}{lrrrrrrrr}
        \toprule
        \multirow{2}{*}[-0.5ex]{Method}
            & \multicolumn{2}{c}{$\mathtt{surgery}$}
            & \multicolumn{2}{c}{$\mathtt{mushroom}$}
            & \multicolumn{2}{c}{$\mathtt{bank}$}
            & \multicolumn{2}{c}{$\mathtt{adult}$} \\
        \cmidrule(lr){2-3} \cmidrule(lr){4-5} \cmidrule(lr){6-7} \cmidrule(lr){8-9}
            & \multicolumn{1}{c}{$\theta = 4$} & \multicolumn{1}{c}{$\theta = 8$}
            & \multicolumn{1}{c}{$\theta = 4$} & \multicolumn{1}{c}{$\theta = 8$}
            & \multicolumn{1}{c}{$\theta = 4$} & \multicolumn{1}{c}{$\theta = 8$}
            & \multicolumn{1}{c}{$\theta = 4$} & \multicolumn{1}{c}{$\theta = 8$} \\
        \midrule
        bAUC-Integer            & 16.71 & 11.02 & 21.75 & 17.33 & 88.31 & 34.40 & 91.16 & 36.27 \\
                              &($\pm$1.44)&($\pm$2.30)&($\pm$1.38)&($\pm$0.57)&($\pm$10.32)&($\pm$9.79)&($\pm$36.84)&($\pm$13.65)\\
        \addlinespace[2mm]
        bAUC-Rounding             & 17.20 & 14.73 & 22.74 & 28.18 & 26.94 & 19.17 & 38.93 & 25.84 \\
                              &($\pm$2.04)&($\pm$1.15)&($\pm$0.92)&($\pm$3.53)&($\pm$?1.40)&($\pm$1.24)&($\pm$?7.33)&($\pm$?2.88)\\
        \addlinespace[2mm]
        $L_1$-Regularization  & 4.25 & 6.78 & 0.26 & 1.43 & 1.10 & 3.12 & 0.35 & 1.84 \\
                              &($\pm$0.22)&($\pm$1.31)&($\pm$0.01)&($\pm$0.07)&($\pm$?0.40)&($\pm$1.02)&($\pm$?0.01)&($\pm$?0.14)\\
        \addlinespace[2mm]
        Elastic-Net           & 3.12 & 7.71 & 0.39 & 1.03 & 0.84 & 3.93 & 1.04 & 5.24 \\
                              &($\pm$0.21)&($\pm$2.43)&($\pm$0.02)&($\pm$0.09)&($\pm$?0.03)&($\pm$0.14)&($\pm$?0.06)&($\pm$?0.13)\\
        \addlinespace[2mm]
        Forward-Selection     & 0.60 & 2.85 & 1.81 & 27.70 & 1.59 & 4.04 & 2.65 & 10.43 \\
                              &($\pm$0.11)&($\pm$0.47)&($\pm$0.12)&($\pm$9.43)&($\pm$?0.05)&($\pm$0.29)&($\pm$?0.03)&($\pm$?0.51)\\
        \addlinespace[2mm]
        Backward-Elimination  & 2.53 & 2.37 & 52.15 & 57.24 & 21.40 & 19.65 & 67.81 & 64.24 \\
                              &($\pm$0.07)&($\pm$0.05)&($\pm$3.61)&($\pm$3.07)&($\pm$?1.50)&($\pm$0.70)&($\pm$?1.52)&($\pm$?1.37)\\
        \bottomrule
        \end{tabular}
    \end{subtable}

    \vspace{3mm}

    \begin{subtable}{\textwidth}
        \centering
        \caption{$\theta = 6$}
        \label{tab:comp_M}
        \begin{tabular}{lrrrrrrrr}
        \toprule
        \multirow{2}{*}[-0.5ex]{Method}
            & \multicolumn{2}{c}{$\mathtt{surgery}$}
            & \multicolumn{2}{c}{$\mathtt{mushroom}$}
            & \multicolumn{2}{c}{$\mathtt{bank}$}
            & \multicolumn{2}{c}{$\mathtt{adult}$} \\
        \cmidrule(lr){2-3} \cmidrule(lr){4-5} \cmidrule(lr){6-7} \cmidrule(lr){8-9}
            & \multicolumn{1}{c}{$M = 1$} & \multicolumn{1}{c}{$M = 2$}
            & \multicolumn{1}{c}{$M = 1$} & \multicolumn{1}{c}{$M = 2$}
            & \multicolumn{1}{c}{$M = 1$} & \multicolumn{1}{c}{$M = 2$}
            & \multicolumn{1}{c}{$M = 1$} & \multicolumn{1}{c}{$M = 2$} \\
        \midrule
        bAUC-Integer            & 10.45 & 12.80 & 20.01 & 19.56 & 38.71 & 91.33 & 42.78 & 112.50 \\
                              &($\pm$1.11)&($\pm$2.10)&($\pm$1.40)&($\pm$1.29)&($\pm$6.74)&($\pm$8.85)&($\pm$16.12)&($\pm$58.17)\\
        \addlinespace[2mm]
        bAUC-Rounding             & 15.28 & 15.74 & 23.95 & 28.18 & 23.76 & 24.29 & 35.73 & 33.26 \\
                              &($\pm$1.61)&($\pm$1.90)&($\pm$1.76)&($\pm$1.84)&($\pm$1.97)&($\pm$1.93)&($\pm$?5.79)&($\pm$?5.81)\\
        \addlinespace[2mm]
        $L_1$-Regularization  & 5.31 & 5.60 & 0.66 & 0.78 & 1.43 & 1.47 & 0.81 & 0.77 \\
                              &($\pm$0.48)&($\pm$0.59)&($\pm$0.01)&($\pm$0.05)&($\pm$0.53)&($\pm$0.52)&($\pm$?0.03)&($\pm$?0.02)\\
        \addlinespace[2mm]
        Elastic-Net           & 3.96 & 3.95 & 0.50 & 0.59 & 2.38 & 2.46 & 1.78 & 1.74 \\
                              &($\pm$0.44)&($\pm$0.46)&($\pm$0.02)&($\pm$0.02)&($\pm$0.10)&($\pm$0.05)&($\pm$?0.04)&($\pm$?0.08)\\
        \addlinespace[2mm]
        Forward-Selection     & 1.87 & 1.89 & 3.09 & 3.56 & 2.83 & 2.70 & 5.31 & 5.13 \\
                              &($\pm$0.73)&($\pm$0.72)&($\pm$0.21)&($\pm$0.10)&($\pm$0.13)&($\pm$0.16)&($\pm$?0.28)&($\pm$?0.29)\\
        \addlinespace[2mm]
        Backward-Elimination  & 2.33 & 2.44 & 51.07 & 60.85 & 21.19 & 21.74 & 71.41 & 69.92 \\
                              &($\pm$0.03)&($\pm$0.08)&($\pm$4.12)&($\pm$3.50)&($\pm$0.79)&($\pm$0.50)&($\pm$?1.37)&($\pm$?2.80)\\
        \bottomrule
        \end{tabular}
    \end{subtable}
\end{table*}

\subsection{Computation times}
Table~\ref{tab:computation} presents the training computation times (in seconds) required for each method. 
Table \ref{tab:comp_theta} shows the effect of varying the maximum number of questions $\theta \in \{4, 8\}$ when the upper bound on regression coefficients was fixed at $M = 2$, whereas Table \ref{tab:comp_M} shows the effect of varying the upper bound on regression coefficients $M \in \{1, 2\}$ when the maximum number of questions was fixed at $\theta = 6$. 

Table \ref{tab:comp_theta} shows that the computation time for the bAUC-Integer method ranged from 10 to 100 s, which was comparable to or up to three times longer than that for the bAUC-Rounding method.
Notably, increasing $\theta$ from 4 to 8 reduced the computation time for the bAUC-Integer method, especially on the large-sized $\mathtt{bank}$ and $\mathtt{adult}$ datasets.
This suggests that relaxing the cardinality constraint in Eq.~\eqref{eq:cardinality} by increasing $\theta$ can reduce the difficulty of the optimization problem.
In contrast, although the $L_1$-Regularization and Elastic-Net methods were very fast, their computation time increased with increasing $\theta$.
The Forward-Selection method significantly increased the computation time with increasing $\theta$, while the Backward-Elimination method took more than 50 s for the $\mathtt{mushroom}$ and $\mathtt{adult}$ datasets.

Table \ref{tab:comp_M} shows that the computation time for the bAUC-Integer method tended to increase from $M=1$ to $M=2$, especially on the large-sized $\mathtt{bank}$ and $\mathtt{adult}$ datasets.
This behavior is consistent with theory: increasing the upper bound on integer coefficients increases computation time by expanding the feasible search space of the optimization problem.
In contrast, since the bAUC-Rounding method does not impose integer constraints on regression coefficients, its computation time was relatively stable (around 15--35 s) regardless of the value of $M$.
Although the computation times for the regularization and stepwise methods were less affected by the value of $M$, the Backward-Elimination method still required more than 50 s for the $\mathtt{mushroom}$ and $\mathtt{adult}$ datasets.

These results highlight the clear trade-off between predictive accuracy and computational efficiency.
Scoring systems for decision-making (e.g., in medicine, criminal justice, and finance) are typically trained offline and are not updated frequently in practice.
For this reason, given the very good predictive accuracy of our bAUC-Integer method, its computation time of less than a few minutes is generally acceptable for practical deployment.

\begin{table*}[!tbh]
\centering
\caption{Score tables generated by our bAUC-Integer method $(M=2,~\theta = 6)$}
\label{tab:risk-scores}

\begin{subtable}{\textwidth}
\centering
\caption{$\mathtt{surgery}$ dataset}
\label{tab:risk_surgery}
\begin{tabular}{l l c}
\toprule
Question & Response & Points \\
\midrule
1. Diabetes mellitus & True & $+1$ \\ [1mm]
2. Smoking           & True & $+1$ \\ [1mm]
3. Diagnosis        & DGN5 & $+2$ \\
& DGN6 & $-1$ \\ [1mm]
4. Forced expiratory volume in 1 second & $\ge 3$ & $-1$ \\ [1mm]
5. Performance status & PRZ2 & $+1$ \\ [1mm]
6. Clinical T (original tumour size) & OC14 & $+2$ \\
\midrule
 \multicolumn{2}{r}{$\textbf{SCORE}=$}\\
\bottomrule
\end{tabular}

\vspace{0.8em}

{\setlength{\tabcolsep}{3pt}
\begin{tabular}{ccccccccccc}
\toprule
\textbf{SCORE} & $-2$ & $-1$ & 0 & 1 & 2 & 3 & 4 & 5 & 6 & 7\\
\midrule
\textbf{PROB.} & 0.8\% & 2.2\% & 5.7\% & 14.0\% & 30.7\% & 54.7\% & 76.6\% & 89.9\% & 96.0\% & 98.5\%\\
\bottomrule
\end{tabular}
}

\end{subtable}

\vspace{2.0em}

\begin{subtable}{\textwidth}
\centering
\caption{$\mathtt{mushroom}$ dataset}
\label{tab:risk_mushroom}

\begin{tabular}{l l c}
\toprule
Question & Response & Points \\
\midrule
1. Gill spacing           & Crowded & $-1$ \\ [1mm]
2. Gill size     & Narrow & $+1$ \\ [1mm]
3. Odor & None & $-1$ \\ [1mm]
4. Stalk root & Club, Cup, Rhizomorphs, or Rooted & $-1$ \\ [1mm]
5. Stalk surface above ring & Smooth & $-1$ \\
& Scaly & $+1$\\
6. Ring number & One & $-1$ \\ [1mm]
\midrule
 \multicolumn{2}{r}{$\textbf{SCORE}=$}\\
\bottomrule
\end{tabular}

\vspace{0.8em}

\begin{tabular}{ccccccccc}
\toprule
\textbf{SCORE} & $-5$ & $-4$ & $-3$ & $-2$ & $-1$ & 0 & 1 & 2 \\
\midrule
\textbf{PROB.} & 4.1\% & 10.4\% & 24.1\% & 46.2\% & 70.0\% & 86.4\% & 94.5\% & 97.9\% \\
\bottomrule
\end{tabular}
\end{subtable}
\end{table*}

\begin{table*}[htbp]\ContinuedFloat
\caption{Score tables generated by our bAUC-Integer method $(M=2,~\theta = 6)$ (continued)}
\begin{subtable}{\textwidth}
\centering
\caption{$\mathtt{bank}$ dataset}
\label{tab:risk_bankmarketing}

\begin{tabular}{l l c}
\toprule
Question & Response & Points \\
\midrule
1. Age & $36\text{--}44$ & $+1$ \\ [1mm]
2. Occupation & Technician & $-1$ \\ [1mm]
3. Education level & Primary & $-1$ \\ [1mm]
4. Contact communication type & Missing & $-2$ \\ [1mm]
5. Last contact day of the month & After the 25th & $+1$ \\ [1mm]
6. Previous campaign outcome & Success & $+1$ \\
& Failure & $-1$ \\
& Missing & $-1$ \\
\midrule
 \multicolumn{2}{r}{$\textbf{SCORE}=$}\\
\bottomrule
\end{tabular}

\vspace{0.8em}

{\setlength{\tabcolsep}{3pt}
\begin{tabular}{ccccccccccc}
\toprule
\textbf{SCORE} & $-6$ & $-5$ & $-4$ & $-3$ & $-2$ & $-1$ & 0 & 1 & 2 & 3\\
\midrule
\textbf{PROB.} & 0.0\% & 0.1\% & 0.5\% & 1.4\% & 3.8\% & 9.7\% & 22.5\% & 44.2\% & 68.2\% & 85.4\% \\
\bottomrule
\end{tabular}
}
\end{subtable}

\vspace{2.0em}
\begin{subtable}{\textwidth}
\centering
\caption{$\mathtt{adult}$ dataset}
\label{tab:risk_adult}

\begin{tabular}{l l c}
\toprule
Question & Response & Points \\
\midrule
1. Age & $\le 30$ & $-1$ \\ [1mm]
2. Workclass           & Self-employed not incorporated & $-1$ \\ [1mm]
3. Education level    & $\ge$ Associate degree – vocational\footnote{This response corresponds to Associate degree – vocational, Associate degree – academic, Bachelors, Masters, Prof-school, or Doctorate.} & $+1$ \\ [1mm]
4. Marital status & Married-civ-spouse & $+2$ \\ [1mm]
5. Capital gain & $\ge \$ 7000$ & $+2$ \\ [1mm]
6. Hours per week & $\le 39$ & $-1$ \\
 & $\ge 41$ & $+1$ \\
\midrule
 \multicolumn{2}{r}{$\textbf{SCORE}=$}\\
\bottomrule
\end{tabular}

\vspace{0.8em}

{\setlength{\tabcolsep}{3pt}
\begin{tabular}{ccccccccccc}
\toprule
\textbf{SCORE} & $-3$ & $-2$ & $-1$ & 0 & 1 & 2 & 3 & 4 & 5 & 6\\
\midrule
\textbf{PROB.} & 0.3\% & 0.8\% & 2.1\% & 5.7\% & 14.2\% & 31.0\% & 55.0\% & 76.9\% & 90.0\% & 96.1\%\\
\bottomrule
\end{tabular}
}
\end{subtable}
\end{table*}

\subsection{Analysis of the generated score tables}

Table \ref{tab:risk-scores} shows the score tables generated by our bAUC-Integer method with the upper bound on regression coefficients $M=2$ and the maximum number of questions $\theta=6$.
As shown in the tables, the estimated probability of a target event occurring increases monotonically with the total score, allowing users to easily predict the event by summing the integer points assigned to each response.

In the $\mathtt{surgery}$ dataset (Table \ref{tab:risk_surgery}) aimed at predicting death within one year after surgery, the assigned scores identified clinically important explanatory factors, such as diabetes, smoking, and original tumour size.
In addition, these factors were assigned positive points, indicating a higher risk of death within one year.
Conversely, a large forced expiratory volume ($\ge 3$) was assigned negative points, which is consistent with the medical finding that better lung function is associated with lower surgical risk.
These observations demonstrate the ability of our method to automatically identify and quantify important clinical indicators.

In the $\mathtt{mushroom}$ dataset (Table~\ref{tab:risk_mushroom}) aimed at identifying poisonous mushrooms, the assigned scores highlight several easily observable morphological features that clearly distinguish high-risk from low-risk mushrooms.
Narrow gill size and a scaly stalk surface above the ring were assigned positive points, indicating that these features are characteristic of poisonous mushrooms.
In contrast, crowded gill spacing, no odor, specific stalk-root types, a smooth stalk surface above the ring, and a single ring were each assigned negative points, lowering the estimated risk of being poisonous.
The resultant risk probability shows a rapid change relative to the score; for example, scores above 0 indicate a greater than 86.4\% probability of being poisonous.

In the $\mathtt{bank}$ dataset (Table~\ref{tab:risk_bankmarketing}) aimed at predicting subscription of a term deposit, the selected variables emphasize the importance of contact information, client profile, and past interactions.
A missing contact communication type was penalized by $-2$ points, indicating that incomplete contact information is strongly associated with a lower probability of subscription. 
Conversely, if the last contact date of the month was after the 25th, or if the previous campaign was successful, $+1$ point was assigned in each case, reflecting favorable timing and positive engagement history.
Clients aged 36--44 were also associated with a higher probability of subscription.
In contrast, clients with primary education and those working as technicians were penalized, suggesting lower responsiveness in these segments. 
The estimated probability of subscription rises from 3.8\% for a score of $-2$ to 68.2\% for a score of $2$. 
This suggests that predicted subscription rates can be significantly increased by improving contact quality and focusing on clients with favorable characteristics and successful past interactions.

In the $\mathtt{adult}$ dataset (Table~\ref{tab:risk_adult}) aimed at predicting high income, the selected variables clearly reflect socioeconomic status associated with high income.
High educational level (at least an associate degree), being married to a civilian spouse, large capital gains ($\ge \$ 7000$), and long working hours ($\ge 41$ hours per week) were all assigned positive points, in line with common economic intuition regarding the determinants of high income.
In contrast, young age ($\le 30$), being self-employed but unincorporated, and short working hours ($\le 39$ hours per week) were assigned negative points.
The resulting score stratifies the population sharply, with the estimated probability of high income increasing from 0.3\% at a score of $-3$ to 96.1\% at a score of $6$.

\section{Conclusion}

Scoring systems are low-dimensional linear classification models with small integer regression coefficients, allowing predictions to be made with simple manual calculations without the need for a calculator.
We proposed an effective MIO framework to construct scoring systems for accurate predictions. 
Specifically, we built on the existing RiskSLIM formulation~\citep{ustun2019learning}, imposed the group sparsity constraint for limiting the number of questions, and maximized bAUC as the tightest concave lower bound on AUC. 
We further reformulated this optimization model as an equivalent MILO problem, which can be handled by standard MIO solvers. 

We conducted comprehensive experiments using real-world datasets for binary classification downloaded from the UC Irvine Machine Learning Repository~\citep{uci_repository}. 
Our method achieved better predictive accuracy than did the baseline methods in three out of the four real-world datasets.
On the other hand, there was a clear trade-off between predictive accuracy and computational efficiency; although our method was often computationally slower than the baseline methods, its computation time was still within a practical range of a few minutes.
We also conducted a detailed analysis of the generated score tables and validated our method based on domain knowledge.

Future research directions include developing efficient algorithms for solving the bAUC maximization problem on large datasets.
While we reduced the solution time for the optimization problem by sampling a small subset of data instances, using all data instances will further improve the predictive performance and robustness of our method.
Another direction for future research is to design an exact solution method that directly maximizes AUC of scoring systems.
Although direct maximization of AUC is known to be an extremely difficult problem~\citep{norton2019maximization}, it will lead to further improvements in the predictive accuracy of scoring systems.

\bibliography{ref}

\end{document}